\documentclass[conference]{IEEEtran}
\IEEEoverridecommandlockouts
\usepackage{cite}

\usepackage{amsmath,amssymb,amsfonts}
\usepackage{amsthm}
\usepackage{mathrsfs}

\usepackage{array}
\usepackage{booktabs}
\usepackage{multirow}
\usepackage{tabularray}

\usepackage{graphicx}
\usepackage{subcaption}

\usepackage{xcolor}
\usepackage{soul}
\definecolor{lightyellow}{RGB}{255,255,204}
\sethlcolor{lightyellow}

\usepackage{textcomp}
\usepackage{url}
\usepackage{verbatim}
\usepackage{manyfoot}
\usepackage[mathlines, switch]{lineno}
\usepackage[justification=centering]{caption}
\usepackage{xspace}

\usepackage{algorithm}
\usepackage{algorithmicx}
\usepackage{algpseudocode}

\usepackage{listings}

\usepackage{stfloats}

\newcommand{\ourpipeline}{CLIFE\xspace}

\def\BibTeX{{\rm B\kern-.05em{\sc i\kern-.025em b}\kern-.08em
    T\kern-.1667em\lower.7ex\hbox{E}\kern-.125emX}}

\begin{document}
%\thanks{This research was partially supported by the Federal Highway Administration (FHWA) Exploratory Advanced Research (EAR) Program under the project titled "Advanced Artificial Intelligence Research for Safety of Vulnerable Road Users", Award No. 693JJ32350028.}

\title{\vspace{0.20in}CLIFE: Camera–LiDAR Fusion Framework for Edge-Deployable Roadside VRU Perception \\
\thanks{This research was partially supported by the Federal Highway Administration under Agreement No. 693JJ32350028. Any opinions, findings, and conclusions or recommendations expressed in this publication are those of the authors and do not necessarily reflect the view of the Federal Highway Administration. \\
\copyright~2026 IEEE. Personal use of this material is permitted.
Permission from IEEE must be obtained for all other uses, in any current
or future media, including reprinting/republishing this material for
advertising or promotional purposes, creating new collective works,
for resale or redistribution to servers or lists, or reuse of any
copyrighted component of this work in other works.}}

% {\footnotesize \textsuperscript{*}Note: Sub-titles are not captured for https://ieeexplore.ieee.org  and
% should not be used}

% \thanks{This research is supported by the Federal Highway Administration (FHWA) Exploratory Advanced Research (EAR) Program, titled "Advanced Artificial Intelligence Research for Equitable Safety of Vulnerable Road Users". Award No.: 693JJ32350028.

% \textsuperscript{1} Tam Bang, Hoang H. Nguyen, Austin Harris, and Mina Sartipi are with the Center for Urban Informatics and Progress, University of Tennessee at Chattanooga.

% \textsuperscript{2} Lei Chech is with the University of Maryland, College Park.

% \textsuperscript{3} Lihao Gou, and Siyang Cao are with the Department of Electrical and Computer Engineering, University of Arizona.}
% }

% \author{\IEEEauthorblockN{Tam Bang\textsuperscript{1}, Hoang H. Nguyen\textsuperscript{1}, Lei Cheng\textsuperscript{2}, Lihao Gou\textsuperscript{3}, Siyang Cao\textsuperscript{3}, Austin Harris\textsuperscript{1}, Mina Sartipi\textsuperscript{1}}
% }

\author{
\IEEEauthorblockN{
Tam Bang\IEEEauthorrefmark{1},
Hoang H. Nguyen\IEEEauthorrefmark{1},
Lei Cheng\IEEEauthorrefmark{2},
Lihao Guo\IEEEauthorrefmark{3},
Siyang Cao\IEEEauthorrefmark{3}
}
\IEEEauthorblockN{
Hussam Abubakr\IEEEauthorrefmark{1},
Tianya Zhang\IEEEauthorrefmark{1},
Austin Harris\IEEEauthorrefmark{1},
Mina Sartipi\IEEEauthorrefmark{1}
}
\IEEEauthorblockA{\IEEEauthorrefmark{1}Center for Urban Informatics and Progress, University of Tennessee at Chattanooga, USA}
\IEEEauthorblockA{\IEEEauthorrefmark{2}University of Maryland, College Park, USA}
\IEEEauthorblockA{\IEEEauthorrefmark{3}University of Arizona, USA}
}

\maketitle

% \begin{strip}
%   \centering
%   \includegraphics[width=\textwidth]{Figure/problems_v5.png}
%   \captionof{figure}{Challenges in roadside VRU perception: (a) camera degradation under low light/glare; (b) limited LiDAR point density affecting VRU detection.}
%   \label{fig:problems}
% \end{strip}

% \begin{figure*}[!htb]
%   \centering
%   \includegraphics[width=0.485\textwidth]{Figure/problems_v5.png}
%   \caption{Challenges in roadside VRU perception: (a) Camera performance degradation due to low light and glare, and (b) limited LiDAR point cloud density affecting VRU detection.}
%   \label{fig:problems}
% \end{figure*}

\begin{abstract}
% Reliable roadside perception of vulnerable road users (VRUs) remains challenging under occlusions, variable lighting, and diverse weather conditions, particularly under strict edge compute and latency constraints. Existing multi-sensor fusion systems rely on cloud or server-grade infrastructure, creating a deployment gap at real-world intersections. We present CLIFE, an edge-native camera–LiDAR fusion framework that integrates targetless online calibration and lightweight late-fusion tracking entirely on a single embedded device, without cloud offloading. CLIFE adaptively refines camera–LiDAR alignment on demand and performs multi-sensor fusion and track association with \(O(N \log N)\) per-frame cost. We deploy CLIFE across 12 signalized intersections in Chattanooga and conduct an in-depth evaluation at a representative intersection using synchronized camera–LiDAR data that spans diverse day and night, as well as weather conditions. Our experiments demonstrate that the fusion architecture substantially enhances the perceptual range and robustness of the individual sensors under varied environmental and traffic conditions. The late-fusion core operates at 53.2 FPS on the Jetson AGX Thor, ensuring high throughput for real-time intersection-scale applications. By centering perception at the edge, CLIFE provides a deployable foundation for downstream safety applications, while reducing bandwidth and calibration overhead for agencies operating multi-intersection corridors.

Reliable roadside perception of vulnerable road users (VRUs) remains challenging under occlusions, variable lighting, and diverse weather conditions, particularly under strict edge-computing and latency constraints. Existing multi-sensor fusion systems rely on cloud or server-grade infrastructure, creating a deployment gap at real-world intersections. We present CLIFE, an edge-native camera–LiDAR fusion framework that integrates targetless online calibration and lightweight late-fusion tracking entirely on a single embedded device, without cloud offloading. CLIFE adaptively refines camera–LiDAR alignment on demand and performs multi-sensor fusion and track association with \(O(N \log N)\) per-frame cost. We deploy CLIFE across 12 signalized intersections in Chattanooga and conduct detailed evaluation at a representative intersection under diverse real-world conditions. Our experiments demonstrate that the fusion architecture substantially enhances the perceptual range and robustness of the individual sensors under varied environmental and traffic conditions. The late-fusion core operates at 53.2 FPS on the Jetson AGX Thor, ensuring high throughput for real-time intersection-scale applications. By centering perception at the edge, CLIFE provides a deployable foundation for downstream safety applications, while reducing bandwidth and calibration overhead for agencies operating multi-intersection corridors.

\end{abstract}


\section{Introduction}
\begin{figure}[ht]
  \centering
  \captionsetup{justification=justified, singlelinecheck=false}
  \includegraphics[width=0.485\textwidth]{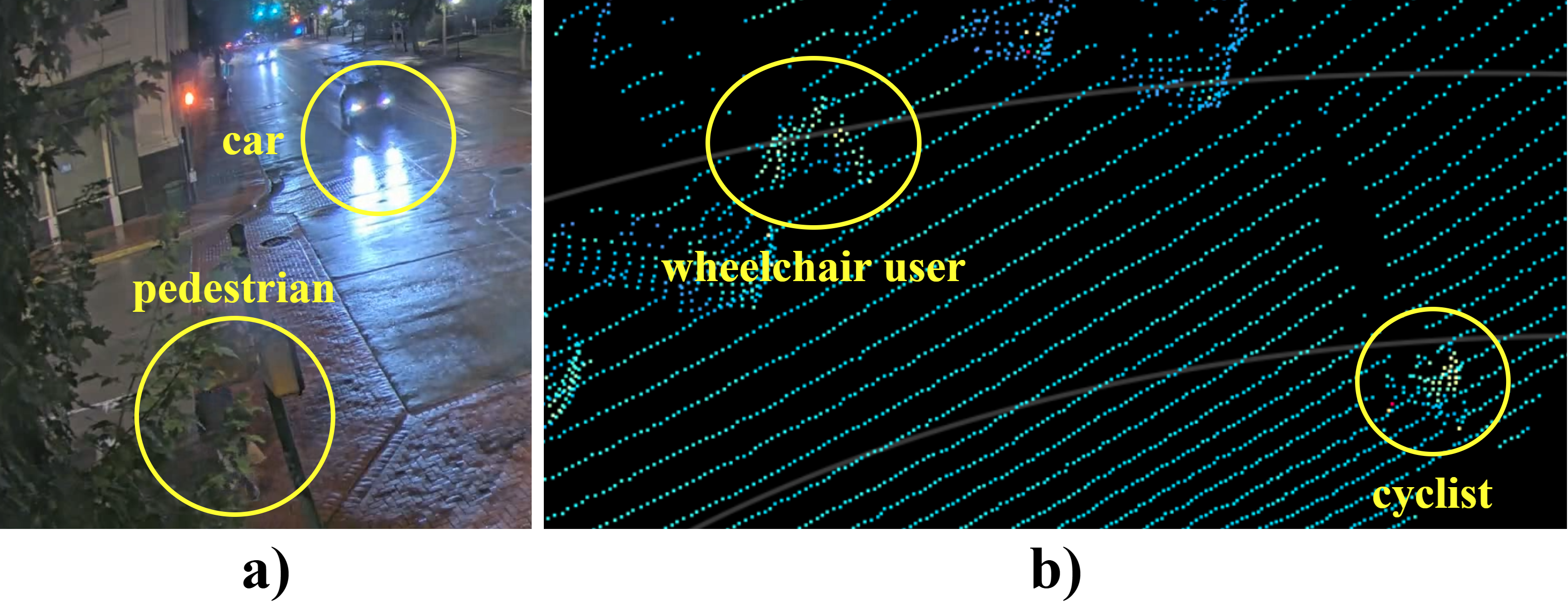}
  \caption{Roadside VRU perception challenges: (a) camera degradation under low light and glare, and (b) sparse LiDAR point clouds.}
  \label{fig:problems}
\end{figure}

Vulnerable Road Users (VRUs) --- pedestrians, cyclists, scooterists, and wheelchair users --- face elevated safety risks in urban traffic environments, where infrastructure often prioritizes vehicular efficiency over their protection \cite{fhwa2023}. Detection at intersections is particularly challenging due to frequent conflicts, occlusions from large vehicles, variable lighting, and diverse weather conditions, with strict requirements on perception-to-decision latency under limited on-device compute. In the United States, VRUs accounted for 21\% of traffic fatalities in 2023 \cite{iihs2023}, underscoring the need for reliable, intersection-scale perception. Fig.~\ref{fig:problems} highlights two representative challenges: (a) camera performance degradation due to low light and glare, causing missed detections, and (b) sparse LiDAR point clouds that degrade detection of small VRUs such as wheelchair users and cyclists.

Despite advances in computer vision, LiDAR, and embedded computing, roadside sensing for VRUs remains limited in real deployments \cite{silva2025vulnerable}. Many existing solutions rely on vehicle-mounted perception~\cite{chen2017mv3d, yin2021mvp}, which depends on uneven fleet adoption and driver behavior. Roadside fusion systems such as InfraDet3D~\cite{zimmer2023infradet3d}, RP3D~\cite{zheng2024rp3d}, and VIPS~\cite{shi2022vips} achieve strong perception performance but require server-grade or centralized backends, increasing deployment cost and bandwidth overhead~\cite{pang2020clocs, liang2019multi}. Single-sensor systems have fundamental limitations: camera-only approaches degrade at night and in glare, while LiDAR-only systems lack appearance cues for fine-grained VRU classification. Both modalities are susceptible to occlusion from large vehicles, though LiDAR's 3D spatial coverage and wider field of view allow it to maintain detections where the camera's 2D perspective is fully blocked. Although extensive literature addresses early, intermediate, and late fusion, most works target on-vehicle autonomy or controlled settings, leaving the gap between research-level perception performance and practical, edge-constrained intersection deployment largely unaddressed~\cite{liu2024multi}.

The practical deployment of multi-sensor roadside systems is further constrained by extrinsic calibration. Sensor fusion algorithms require precise spatial alignment, where extrinsic pose errors or rotation misalignment induce multi-pixel reprojection errors that cascade into false associations and identity fragmentation, particularly for VRUs at long ranges~\cite{an2024survey}. Conventional calibration workflows rely on manual measurements, checkerboard targets, or specialized rigs, requiring skilled technicians, lane closures, and periodic recalibration following environmental perturbations such as thermal expansion or wind-induced mount drift~\cite{li2023automatic}.

Motivated by these deployment constraints, we present CLIFE, a \underline{C}amera–\underline{Li}DAR \underline{F}usion framework for \underline{E}dge-Deployable VRU roadside perception. CLIFE integrates targetless online calibration and lightweight late-fusion tracking into a fully on-device pipeline, running end-to-end on a single embedded device without cloud offloading. The architecture operates as an efficient node with low computational overhead, enabling a single device to serve multiple camera–LiDAR streams and scale to distributed per-intersection deployment. 

To demonstrate the real-world viability and robustness of our approach, we validate CLIFE across a live testbed of 12 signalized intersections in Chattanooga, TN, with in-depth evaluation at a representative intersection under diverse traffic, weather, and lighting conditions. Our contributions are threefold: 

\begin{itemize} 

    \item \textbf{Edge-deployable end-to-end pipeline:} A fully on-device modular pipeline covering camera perception, targetless online calibration, and late-fusion tracking on a single NVIDIA Jetson, without cloud offloading. Each module can be independently trained or updated --- either on the cloud or directly on the edge --- and deployed without re-engineering the full stack.
    
    \item \textbf{Automatic online calibration and lightweight fusion:} Calibration is triggered on demand without manual intervention or calibration targets, seamlessly resuming normal operation upon completion. The fusion core operates with $O(N \log N)$ per-frame complexity, sustaining 53.2 FPS on the Jetson AGX Thor.
    
    \item \textbf{Real-world deployment and evaluation:} Field deployment across 12 signalized intersections, with quantitative evaluation spanning sunny, cloudy, and light-rain conditions, and qualitative results under heavy rain and nighttime glare, demonstrating strong improvements over camera-only baselines and competitive performance across diverse conditions.
    
\end{itemize}

\section{Related Work}

Approaches to multi-modal fusion can be broadly categorized by the stage at which sensor data is combined. Early fusion methods, exemplified by MV3D~\cite{chen2017mv3d} and Liang et al.~\cite{liu2023real}, integrate raw sensor data at the input level, tightly coupling geometry and appearance. Intermediate fusion frameworks, such as BEVFusion~\cite{liu2022bevfusion} and deep structural models~\cite{an2022deep}, instead learn modality-specific features before merging them within a shared network. While these two categories differ in where fusion occurs, both tightly couple sensor streams and demand substantial computational resources, limiting their applicability to embedded edge deployments.

Late fusion combines higher-level outputs from independent sensor pipelines, offering modularity and computational efficiency. CLOCs~\cite{pang2020clocs} and Fast-CLOCs~\cite{pang2022fast} associate camera and LiDAR object candidates before non-maximum suppression to refine 3D detections in autonomous driving contexts. At the decision level, track-based fusion methods merge outputs from camera, LiDAR, and radar trackers in multi-sensor traffic monitoring systems, providing robustness to single-sensor failures at modest computational cost. However, these methods are primarily designed for vehicle-mounted or server-backed deployments and assume offline, manually maintained extrinsic calibration --- a process requiring skilled technicians and periodic lane closures that is impractical for long-term fixed roadside infrastructure~\cite{liu2024multi}.

Roadside fusion systems have received growing attention for intersection-scale perception. InfraDet3D~\cite{zimmer2023infradet3d} integrates multiple LiDARs with cameras mounted on overhead infrastructure to extend 3D detection range at intersections. RP3D~\cite{zheng2024rp3d} fuses multi-view cameras and LiDARs for roadside 3D perception at complex urban junctions. Infrastructure-assisted camera--LiDAR networks~\cite{yao2024infrastructure} and VIPS~\cite{shi2022vips} further explore infrastructure-based perception using centralized or server-grade backends for cooperative autonomous driving applications. While these systems demonstrate strong detection performance, they rely on server-grade backends and offline, manually maintained calibration pipelines. To the best of our knowledge, no prior work has concurrently addressed continuous VRU tracking under edge-computing constraints while simultaneously supporting on-demand, targetless recalibration for long-term roadside deployment.

CLIFE is designed to fill this gap through a cohesive, fully on-device pipeline that integrates detection, calibration, and late-fusion tracking for roadside VRU perception. Two key characteristics distinguish our approach. First, it incorporates on-demand targetless calibration, eliminating manual intervention and recalibration overhead. Second, its lightweight, end-to-end fusion framework runs entirely on a single embedded Jetson, maintaining reliable VRU perception through single-sensor failures, occlusions, and varied lighting and weather conditions. This combination of capabilities has not been addressed by prior roadside fusion systems.
\section{Methodology}

\begin{figure}[h] 
\centering   
\captionsetup{justification=justified, singlelinecheck=false}
\includegraphics[width=0.48\textwidth]{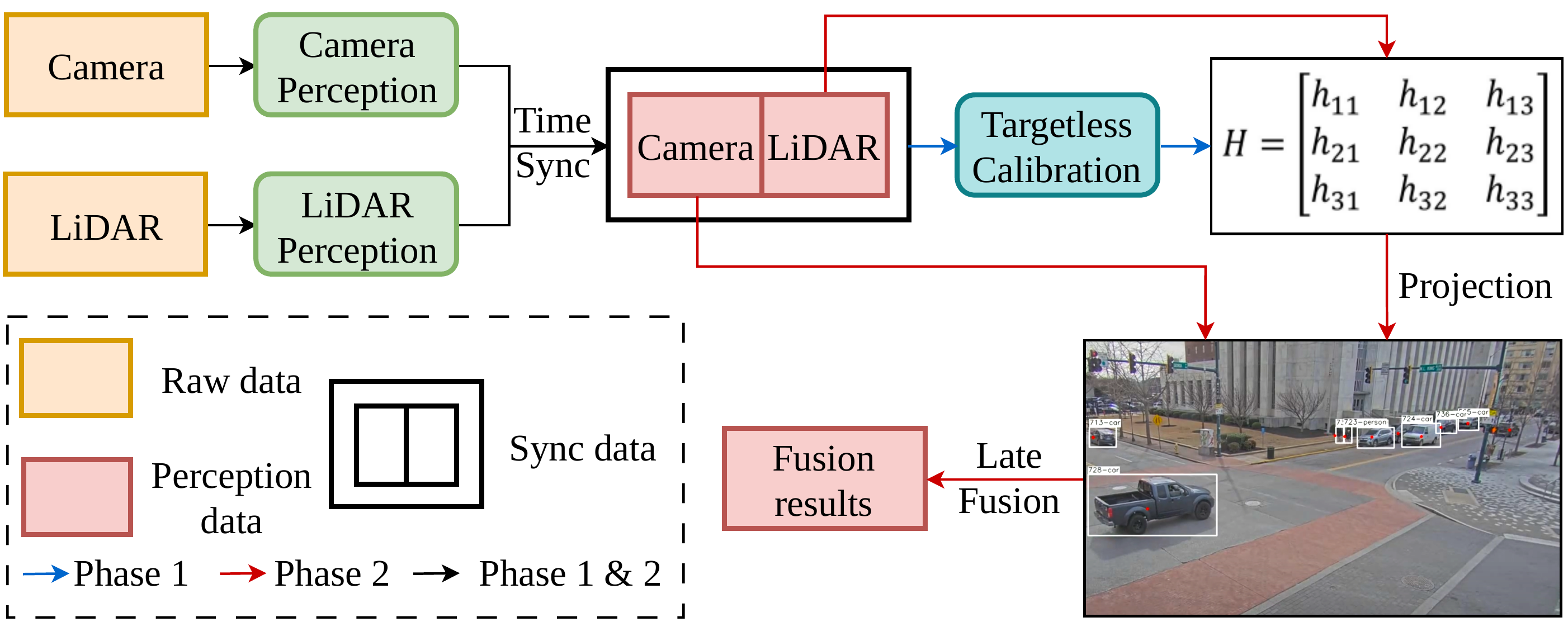}
\caption{Overview of the \ourpipeline architecture. The framework operates in two phases: (1) targetless camera--LiDAR calibration estimates a ground-plane homography, and (2) late fusion associates, and tracks detections across both modalities.} 
\label{fig:Architecture} 
\end{figure}

We present \ourpipeline, a real-time, edge-native framework for roadside VRU perception at urban intersections. The system runs end-to-end on embedded hardware, performing targetless calibration and compute-bounded late fusion entirely on a single edge device. As illustrated in Fig.~\ref{fig:Architecture}, \ourpipeline comprises two phases, both operating on per-frame perception outputs from camera and LiDAR modules. 

\textbf{Phase~1 (Camera--LiDAR Calibration)} employs a targetless, on-device routine that estimates a ground-plane homography $H\in\mathbb{R}^{3\times 3}$ mapping road-plane coordinates to image coordinates. This homography is adaptively updated on demand (e.g., after mount drift or manual camera adjustment) to maintain accurate cross-sensor alignment throughout deployment. \textbf{Phase~2 (Late Fusion)} uses this alignment to project LiDAR detections into the image frame, performs cross-sensor association and attribute fusion (class, confidence, position, velocity), and maintains online multi-object tracks. This design exploits the complementary strengths of image semantics and LiDAR geometry while maintaining bounded per-frame cost for reliable, real-time VRU perception on an edge device.

\subsection{Sensor Perception Module}

The sensor perception module runs two parallel pipelines (camera and LiDAR), each with:
\begin{enumerate}
  \item \textbf{Detection \& classification:} a real-time detector localizes traffic participants and assigns VRU classes.
  \item \textbf{Online tracking:} a multi-object tracker associates detections across frames to maintain stable IDs.
\end{enumerate}

For each \emph{camera frame} with timestamp $\tau_{\mathrm{C}}$, the camera pipeline produces
\begin{equation}
    \Phi_{\mathrm{C}}
    = \bigl(\mathrm{frame},\,\tau_{\mathrm{C}},\, \{obj_{C}^{i} \}_{i=1}^{N}\bigr),\;
\end{equation}
where: \(obj_{C}^{i} = (\mathrm{id},x,y,w,h,\mathrm{cls},\mathrm{conf},t_{init},\mathrm{aux})\).
\\

For each \emph{LiDAR sweep} with timestamp $\tau_{\mathrm{L}}$, the LiDAR pipeline produces
\begin{equation}
    \Phi_{\mathrm{L}}
    = \bigl(\mathrm{pc},\,\tau_{\mathrm{L}},\, \{obj_{L}^{i} \}_{i=1}^{M}\bigr),
\end{equation}
where: \( obj_{L}^{i} = (\mathrm{id},x,y,z,l,w,h,\mathrm{cls},\mathrm{conf},t_{init},\mathrm{aux})\).

\begin{table}[ht]
\captionsetup{justification=justified, singlelinecheck=false}
\centering
\footnotesize
\caption{Compact notation for perception outputs.}
\setlength{\tabcolsep}{4pt}
\renewcommand{\arraystretch}{1.02}
\begin{tabular}{l|l}
\toprule
\textbf{Symbol} & \textbf{Description} \\
\midrule
$\mathrm{frame}$, $\mathrm{pc}$ & Raw RGB image and raw point cloud \\
$\tau_{\mathrm{C}}$, $\tau_{\mathrm{L}}$ & Timestamp of camera frame/LiDAR sweep\\
$\mathrm{id}$, $\mathrm{cls}$, $\mathrm{conf}$  & Track identifier, class label and confidence score \\
$t_{\mathrm{init}}$ & Track-creation time \\
$(x,y,w,h)$ & 2D bounding box (center, size) [px] \\
$(x,y,z,\ell,w,h)$ & 3D bounding box (center, size) [m] \\
$N$, $M$ & Number of objects in camera frame/LiDAR sweep \\
$\mathrm{aux}$ & Auxiliary attributes (sensor-specific) \\
\bottomrule
\end{tabular}
\label{tab:perception-fields}
\end{table}

Table~\ref{tab:perception-fields} summarizes the notation for camera–LiDAR perception outputs. Image-plane terms are in pixels while LiDAR-frame terms are in meters. We use $\mathrm{aux}$ to denote \emph{sensor-specific auxiliary attributes} (e.g., velocity, heading) that are preserved to support auditability and potential downstream applications. We retain the raw \texttt{frame} and point cloud \texttt{pc} in short sliding buffers, where they are consumed by the targetless calibration module. Once alignment is established, the core late-fusion logic operates solely on the structured detection outputs without requiring raw sensor data. Per-timestamp outputs $\Phi_{\mathrm{C}}$ and $\Phi_{\mathrm{L}}$ are buffered and paired into synchronized pairs $\mathcal{S} = \{(\Phi_{\mathrm{C}}, \Phi_{\mathrm{L}})\}$ before entering calibration and fusion.

\subsection{Phase 1: Camera--LiDAR Calibration}

Precise spatial alignment between the camera and LiDAR coordinate frames is essential for reliable sensor fusion. Without proper calibration, projection errors lead to false associations and degraded tracking performance. CLIFE addresses this through a fully automatic, targetless online calibration module deployable on edge devices. Given a batch of time-synchronized camera--LiDAR pairs $\mathcal{S}$, the module estimates a ground-plane homography $H \in \mathbb{R}^{3\times 3}$ by matching cross-modal detections using spatial, appearance, and semantic cues, followed by iterative geometric refinement to minimize reprojection error.

\begin{figure}[h]
\captionsetup{justification=justified, singlelinecheck=false}
\centering
\includegraphics[width=0.4\textwidth]{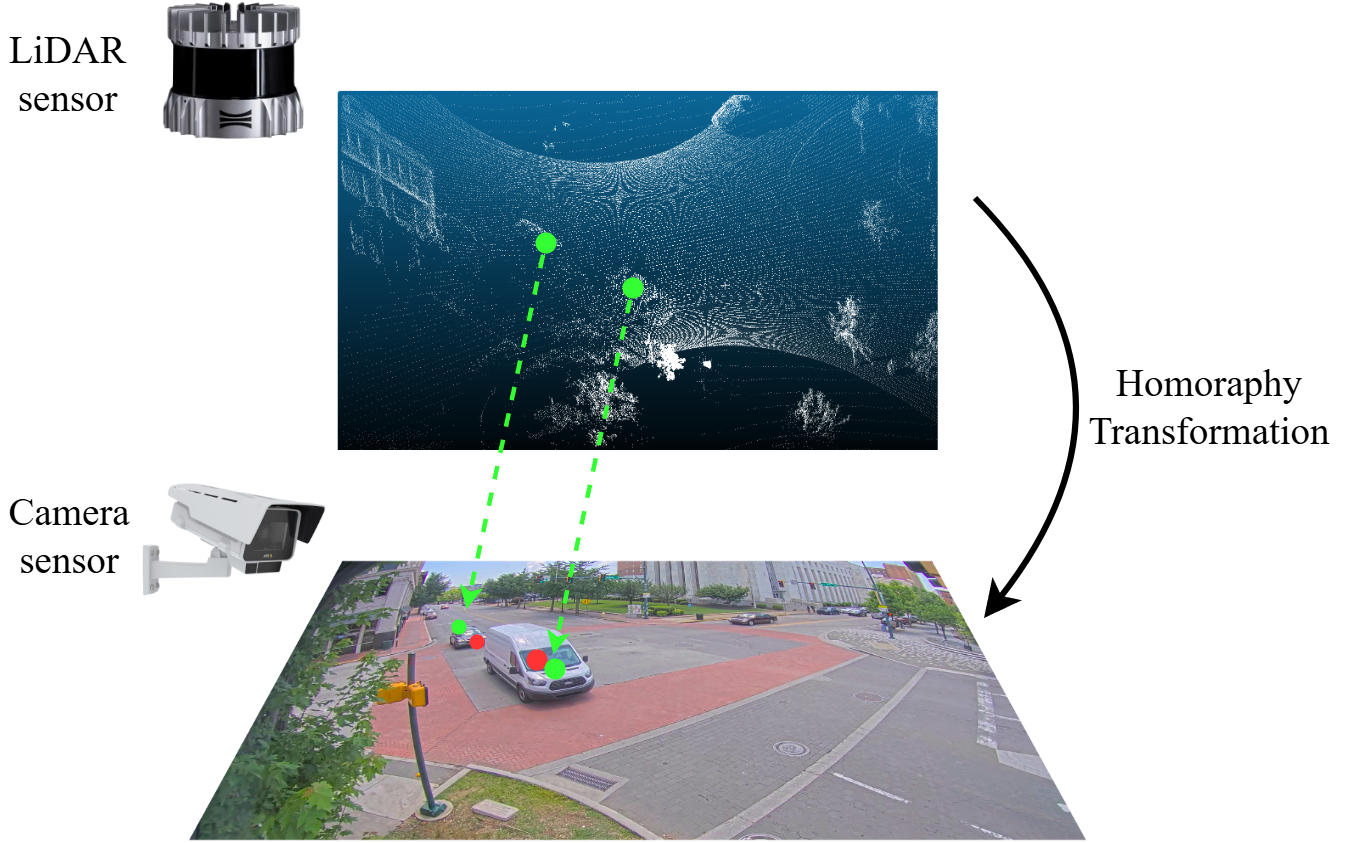}
\caption{Illustration of homography transformation. With accurate calibration, the ground-plane points $(x, y)$, projected via $H$, are aligned with their corresponding object centers in the image.}
\label{fig:Homo_trans}
\end{figure}

Fig.~\ref{fig:Homo_trans} illustrates the resulting alignment, where LiDAR ground-plane points $(x, y)$ are accurately projected to image coordinates via the estimated $H$. This targetless approach eliminates manual setup and enables rapid deployment across diverse intersection geometries, remaining robust to environmental perturbations such as mount drift or thermal expansion.

\subsection{Phase 2: Late Fusion}

Following calibration, \ourpipeline integrates camera and LiDAR perception streams via a lightweight, bounded-cost pipeline. Camera detections provide rich semantics for reliable VRU classification, while LiDAR supplies precise 3D localization and velocity. Fusing these complementary cues preserves robustness under low light, precipitation, partial occlusions, and single-sensor dropouts. Late fusion proceeds in two stages.

\paragraph{Multi-sensor fusion}
Using the calibrated homography \(H\), LiDAR detections are projected onto the image plane to enable matching in a common 2D frame. Matching is performed separately for each category, such as VRUs and vehicles. For each detection, a KD-tree retrieves candidate matches within a search radius \(r_s\) (in pixels), followed by greedy one-to-one assignment that minimizes image-plane Euclidean distance. The full procedure is summarized in Algorithm~\ref{alg:late_fusion}.

\begin{algorithm}[h]
\small
\caption{\small{Distance-based Late Fusion with Track Association}}
\label{alg:late_fusion}
\begin{algorithmic}[1]
\Require Synchronized pair $\mathcal{S}$, homography $H$, projection $\pi_H$, radius search $r_{\mathrm{s}}$, $\tau_{\text{miss}}$, $N_{\text{consec}}$, K
\Ensure Fused $\mathtt{F}$, camera-only $\mathtt{C}$, LiDAR-only $\mathtt{L}$
\State $\mathtt{F}\!\gets\!\varnothing,\; \mathtt{C}\!\gets\!\varnothing,\; \mathtt{L}\!\gets\!\varnothing$
\State \textbf{Project LiDAR:} $\Psi_{\mathrm{L}} \gets \{(\ell,\ \hat{\mathbf{p}}=\pi_H(\text{center}(\ell))) : \ell \in \Phi_{\mathrm{L}}\}$
\For{$k \in \{\mathrm{VRU},\mathrm{vehicle}\}$} \Comment{process per category}
  \State $C_k \gets \{c \in \Phi_{\mathrm{C}} : \text{cls}(c)=k\}$,\quad $L_k \gets \{(\ell,\hat{\mathbf{p}})\in\Psi_{\mathrm{L}} : \text{cls}(\ell)=k\}$
  \If{$C_k=\varnothing$} \State $\mathtt{L}\gets \mathtt{L}\cup\{\ell:(\ell,\hat{\mathbf{p}})\in L_k\}$; \textbf{continue} \EndIf
  \If{$L_k=\varnothing$} \State $\mathtt{C}\gets \mathtt{C}\cup C_k$; \textbf{continue} \EndIf
  \State \textbf{KD-tree build} on $\{\hat{\mathbf{p}}:(\ell,\hat{\mathbf{p}})\in L_k\}$
  \State \textbf{Radius search:} for each $c\!\in\!C_k$, get candidates within $r_s$
  \State \textbf{Greedy-Assign:} one-to-one matches $M_k$ minimizing image-plane distance
  \For{each $(c,\ell)\in M_k$} 
  \State $\mathtt{F}\!\gets\!\mathtt{F}\cup\{\texttt{Fuse}(c,\ell)\}$ \Comment{Attribute fusion} \EndFor
  \State \textbf{Pass-through:} add unmatched $c\!\in\!C_k$ to $\mathtt{C}$ and unmatched $\ell\!\in\!L_k$ to $\mathtt{L}$
\EndFor \\
\Return $\mathtt{F},\ \mathtt{C},\ \mathtt{L}$
\end{algorithmic}
\begin{flushleft}
A FIFO buffer over the last $K$ frames accumulates $\mathtt{F}$, $\mathtt{C}$, and $\mathtt{L}$ for track association (miss counters, ID policy).
\end{flushleft}
\vspace{-5pt}
\end{algorithm}

For each matched pair, attributes are combined according to sensor strengths, ties favor the camera due to superior semantic classification. Class and confidence are taken from the higher-confidence source. Position is retained from the camera frame since fusion operates in image coordinates, while velocity is derived from LiDAR for accurate 3D motion estimation. Unmatched detections are propagated as camera-only or LiDAR-only hypotheses to preserve recall during single-sensor dropouts. Auxiliary fields are forwarded unchanged.

\paragraph{Multi-sensor track association}
The tracker maintains a FIFO buffer over the last $K$ frames, accumulating fused ($\mathtt{F}$), camera-only ($\mathtt{C}$), and LiDAR-only ($\mathtt{L}$) detections to support track association and identity management. At each frame, the tracker ingests new detections and maintains a single external track ID while preserving sensor IDs as auxiliary fields. For fused tracks, the external ID is initialized from earliest track-creation time ($t_{init}$), with ties preferring LiDAR due to its wider field of view. Tracks persist through single-sensor misses via a miss counter that increments when detections are absent. Track termination occurs when this counter exceeds $\tau_{\text{miss}}$ or both sensors fail to report the object for $N_{\text{consec}}$ consecutive frames, ensuring identity consistency and fault tolerance.

The overall pipeline has per-frame complexity $O(N\log N)$ (dominated by KD-tree queries), enabling real-time, low-latency operation on embedded edge hardware while exploiting the complementary strengths of camera semantics and LiDAR geometry for robust VRU perception. Each fused track contains the merged attributes (2D box, position, class, confidence, velocity) alongside the complete sensor metadata from both modalities for downstream analytics.

\section{Experiments}
\subsection{Experimental Setup}
\begin{figure}[htbp]
\centering
\captionsetup{justification=justified, singlelinecheck=false}
\includegraphics[width=0.47\textwidth]{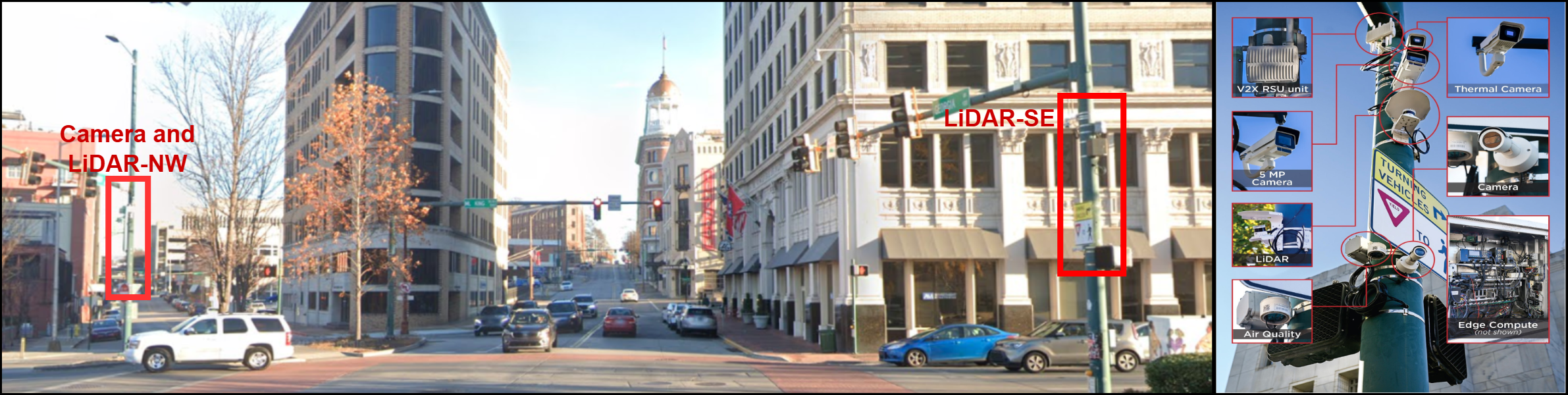}
\caption{Street-view of the sensor setup.}
\label{fig:street_view}
\end{figure}

We deploy the system across 12 signalized intersections in Chattanooga, TN, with in-depth evaluation conducted at the Georgia Avenue and M.L.K.\ Boulevard intersection (Fig.~\ref{fig:street_view}), selected for its frequent and diverse VRU activity. All modules run on a single NVIDIA Jetson AGX Thor in maximum performance mode.

Each intersection is equipped with an Ouster BlueCity backend~\cite{ousterbluecity}, a co-located edge system that runs on-site, time-aligns and merges point clouds from the northwest (NW) and southeast (SE) LiDAR units, and delivers structured perception outputs via API at 10\,Hz. We leverage this existing field infrastructure directly, with CLIFE ingesting the BlueCity perception stream alongside a northwest (NW) camera stream processed locally on the Jetson. Camera and LiDAR perception are synchronized at the LiDAR's native 10\,Hz rate before being passed sequentially to the calibration and late fusion stages. Table~\ref{tab:devices} details the specific hardware configurations.

\begin{table}[ht]
\captionsetup{justification=justified, singlelinecheck=false}
\centering
\scriptsize
\caption{Devices and configurations at an intersection.}
\label{tab:devices}
\setlength{\tabcolsep}{5pt}
\renewcommand{\arraystretch}{1.05}
\begin{tabular}{l|p{60mm}}
\toprule
\textbf{Devices} & \textbf{Configurations} \\
\midrule
Two cameras &
AXIS P1377-LE; \(1920\times1080\) @ 30\,FPS (RTSP); fixed focus; mount height \(\sim\)5\,m; tilt \(20^\circ\) down; FoV covering both crosswalk legs and turn bays. \\
\midrule
Two LiDAR units &
Ouster OS1-128; 128 channels; 10\,Hz; vertical FoV \(\sim45^\circ\); range up to \(\sim200\) m (vendor spec); rigid pole mounts \(\sim\)5\,m at NW and SE; sweeps time-aligned and merged by BlueCity backend. \\
\midrule
One edge device & NVIDIA Jetson AGX Thor; 40W to 130W; housed in a roadside NEMA cabinet with 120\,VAC mains.\\
\bottomrule
\end{tabular}
\end{table}

%--------------------------------------------------------------------------

\subsection{Datasets}
\begin{figure}[h!]
\captionsetup{justification=justified, singlelinecheck=false}
  \centering
  \includegraphics[width=\linewidth]{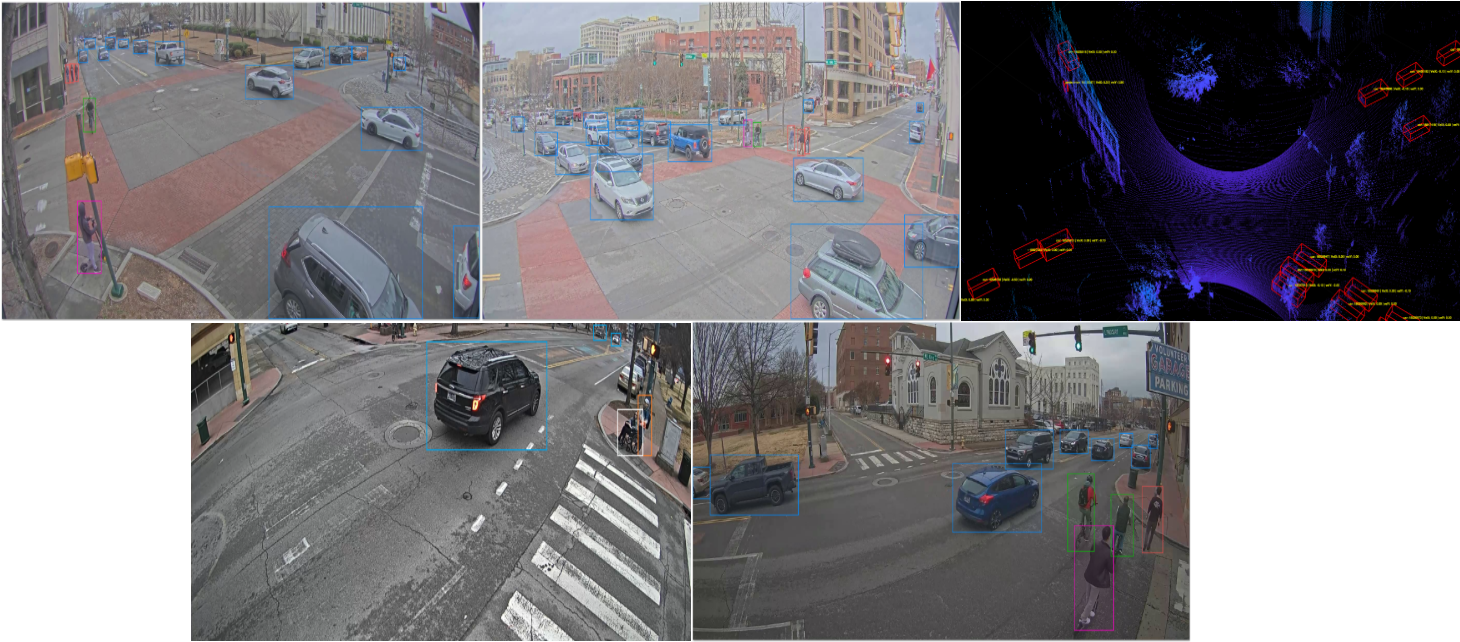}
  \caption{Sample images from the multi-sensor VRU dataset.} 
  \label{fig:dataset_sample}
\end{figure}

To train and evaluate our system, we collected and annotated two custom roadside datasets from intersections in Chattanooga, TN (examples are shown in Fig.~\ref{fig:dataset_sample}).

\paragraph{Camera Perception Dataset} To fine-tune our camera perception model, we curated a dataset of 25{,}270 camera-only frames from multiple viewpoints at the Georgia \& M.L.K. and Lindsay \& M.L.K. intersections. This dataset features scripted scenarios with diverse VRU interactions, from individuals to groups at crosswalks, ensuring representative coverage of infrequent VRU classes for robust model training.

\paragraph{Synchronized Camera--LiDAR Dataset} For end-to-end evaluation, we collected and annotated 9{,}000 time-synchronized camera--LiDAR frame pairs under diverse real-world conditions, evenly distributed across three weather conditions (3{,}000 frames each: sunny, cloudy, and light rain without lens occlusion). More extreme conditions, such as heavy rain and nighttime, were also collected but reserved for qualitative experiments due to the challenges in establishing reliable ground-truth annotation under these conditions.

\paragraph{Annotation Methodology} Annotation followed a two-stage process. Camera frames were first auto-labeled by an off-the-shelf detector and then manually refined in CVAT~\cite{CVAT_ai_Corporation_Computer_Vision_Annotation_2023}, with human review prioritized for rare VRU classes (scooterists, wheelchair users). LiDAR objects were initialized from BlueCity outputs and manually corrected using SUSTechPOINTS~\cite{li2020sustech}. Camera frames were downsampled to 10\,FPS to match the LiDAR's 10\,Hz sweeps, and labeling effort prioritized VRU-rich intervals.

The BlueCity software is limited to a fixed class set (pedestrian, car, truck, bicycle, bus, trailer) without model customization, and sparse point clouds make fine-grained VRU differentiation (e.g., scooterists, wheelchair users) unreliable even with manual annotation. Consequently, camera annotations serve as the primary semantic ground truth for fine-grained VRU classes, while LiDAR contributes complementary geometric context for shared categories.

%--------------------------------------------------------------------------

\subsection{Evaluation Metrics}
We evaluate our system using standard metrics for detection, tracking, and calibration. For object detection, we measure precision, recall, and mean Average Precision (mAP), reported at both mAP@0.5 and the COCO-style mAP@0.5:0.95. For multi-object tracking, we use  Multiple Object Tracking Accuracy (MOTA), and the IDF1 score. Finally, calibration accuracy is quantified by the Average Euclidean Distance (AED) in pixels, measured between \(H\)-projected LiDAR ground points and their matched camera object center, averaged across all synchronized pairs.

\subsection{Experimental Results}

We benchmarked each module on an NVIDIA Jetson AGX Thor at its maximum power profile. We employ two key optimizations to minimize latency: (i) all neural network models are compiled with NVIDIA TensorRT~\cite{nvidiatensorrt}, and (ii) during late fusion, we avoid full-frame warping by transforming only detection coordinates between the distorted and rectified image spaces.

\subsubsection{Camera Perception}

We fine-tune YOLOv11 \cite{khanam2024yolov11} at 1280$\times$1280 resolution to better capture small VRU instances on 25{,}270 camera-only frames using an 80:10:10 train/validation/test split. Table~\ref{tab:yolo_precision_speed} summarizes detection accuracy and throughput across YOLOv11 variants. YOLOv11-m with FP16 at 1280$\times$1280 achieves near-top mAP@0.5:0.95 while offering substantially higher FPS than the -l variant, making it suitable for real-time edge deployment.

\begin{table}[h!]
\captionsetup{justification=justified, singlelinecheck=false}
\centering
\scriptsize
\caption{YOLOv11 mAP@0.5:0.95 and FPS across model variants, input resolutions, and FP16/FP32}
\label{tab:yolo_precision_speed}
\setlength{\tabcolsep}{6pt}
\begin{tabular}{c c c c c c}
\toprule
\multirow{2}{*}{Variant} & \multirow{2}{*}{Input Size} & \multicolumn{2}{c}{FP16} & \multicolumn{2}{c}{FP32} \\
\cmidrule(lr){3-4}\cmidrule(lr){5-6}
 & & mAP@0.5:0.95 & FPS & mAP@0.5:0.95 & FPS \\
\midrule
-s & $640$   & 0.5682& 87.3  & 0.5706 & 80.7 \\
-s & $1280$  & 0.7527 & 46.5 & 0.7566 & 36.2 \\
\midrule
-m & $640$   & 0.6574 & 78.4 & 0.6605 & 61.5 \\
-m & $1280$  & 0.7786 & 39.4 & 0.7815 & 23.7 \\
\midrule
-l & $640$   & 0.6728 & 74.5 & 0.6770 & 55.7 \\
-l & $1280$  & 0.7789 & 33.6 & 0.7838 & 21.6 \\
\bottomrule
\end{tabular}
\end{table}

Table~\ref{tab:class_accuracy} presents per-class detection accuracy, demonstrating robust performance for all VRU categories including challenging classes such as scooterists and wheelchair users.

\begin{table}[htbp]
\captionsetup{justification=justified, singlelinecheck=false}
\centering
\small
\caption{Per-class AP of YOLOv11-m at 1280$\times$1280.}
\label{tab:class_accuracy}
\begin{tabular}{l|c|c}
\toprule
\textbf{Class} & \textbf{Per-class AP@0.5:0.95} & \textbf{Instances} \\
\midrule
Car        & 0.8821 & 21,540 \\
Truck      & 0.8609 & 3,260 \\
Bus        & 0.8715 & 549 \\
Pedestrian & 0.7797 & 18,750 \\
Cyclist    & 0.7836 & 15,156 \\
Scooterist & 0.7282 & 5,934 \\
Wheelchair user & 0.7543 & 5,351 \\
\bottomrule
\end{tabular}
\end{table}

For tracking, we compare ByteTrack \cite{zhang2022bytetrack} and BoT-SORT \cite{aharon2022bot} using the YOLOv11-m. ByteTrack attains comparable identity accuracy to BoT-SORT on the Camera Perception Dataset (IDF1 = 64.3 vs.\ 64.7) while delivering 37\% higher throughput. We therefore adopt \emph{FP16 YOLOv11-m at 1280$\times$1280} together with \emph{ByteTrack} as the default configuration for all subsequent experiments.

\subsubsection{Calibration}
We use CalibRefine~\cite{cheng2025calibrefine}, a targetless, on-device calibration module that estimates the camera--LiDAR homography by matching cross-modal detections through learned spatial, appearance, and semantic cues, followed by iterative geometric refinement. The module's cross-modal matching network was initialized from a pre-trained checkpoint and fine-tuned on our intersection-specific calibration subset to adapt to the roadside sensor geometry and mounting configuration, achieving 94.98\% cross-modal pairing accuracy.

% During initial deployment, we observed peripheral misalignment between camera and LiDAR projections on the original distorted image, as shown in Fig.~\ref{fig:calib_projection} (left). This misalignment stems from lens distortion, which displaces object centers in the image plane and degrades homography accuracy near the field-of-view boundaries. To address this, we apply the calibration module on rectified images instead. Since the learned matching features — spatial positions, appearance embeddings, and semantic labels — are largely preserved after undistortion, the fine-tuned model transfers directly without retraining. Recalibrating on rectified images resolves the peripheral misalignment and achieves precise alignment across the full field of view, as shown in Fig.~\ref{fig:calib_projection} (right).

\begin{figure}[h!]
  \centering
  \captionsetup{justification=justified, singlelinecheck=false}
  \includegraphics[width=\linewidth]{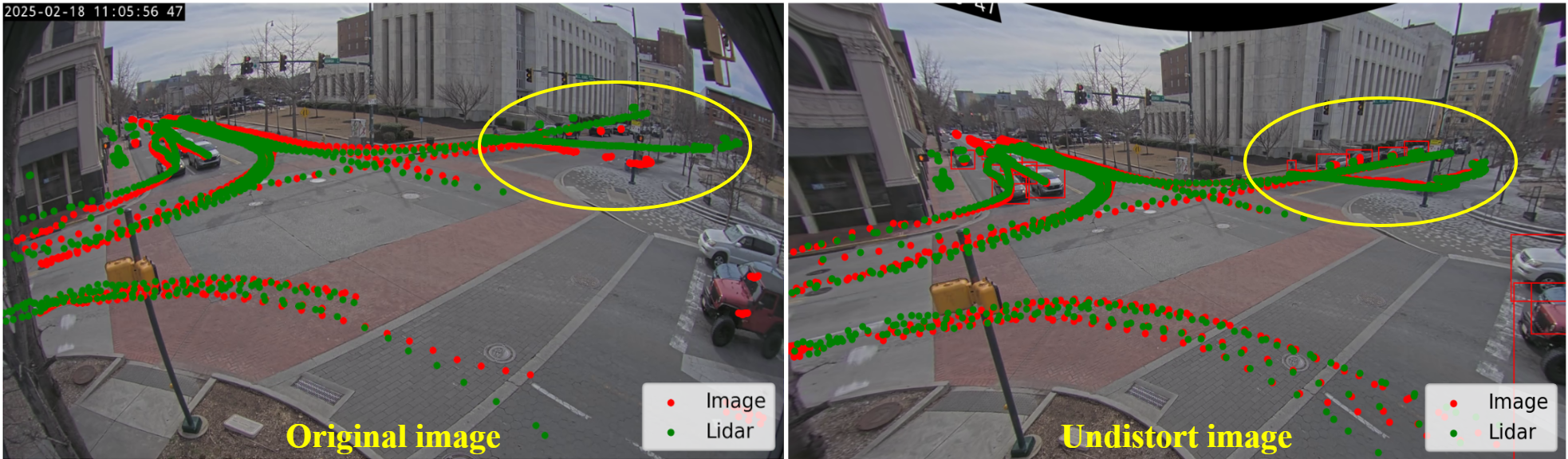}
    \caption{LiDAR ground-plane projections overlaid on the original distorted image (left) and the rectified image (right). The yellow ellipse highlights the peripheral region where lens distortion causes misalignment between camera (red) and LiDAR (green) projections, which is resolved after rectification.}
  \label{fig:calib_projection}
\end{figure}

During initial deployment, we observed peripheral misalignment between camera and LiDAR projections on the original distorted image (Fig.~\ref{fig:calib_projection}, left), caused by lens distortion near the field-of-view boundaries. We therefore perform calibration on rectified images, where the learned matching features—spatial positions, appearance embeddings, and semantic labels—are largely preserved after undistortion, the model transfers without retraining. As shown in Fig.~\ref{fig:calib_projection} (right), recalibration on rectified images resolves the misalignment and restores accurate alignment across the full field of view.

The recalibration on rectified images converges to a final AED of 16.02\,px after approximately 30 minutes of on-site calibration. The module runs on demand as an off-path process, triggered remotely via API to correct sensor mount drift or misalignment without interrupting the primary real-time pipeline.

\subsubsection{End-to-End Tracking Performance}
We evaluate end-to-end tracking performance for three configurations (camera-only, LiDAR-only, and our proposed fusion approach) using a common fusion setup $(r_{\mathrm{s}} = 100,\, \tau_{\text{miss}} = 10,\, N_{\text{consec}} = 5,\, K = 30)$. For fair comparison, all metrics are computed within the NW camera field-of-view mask. The quantitative results in Table~\ref{tab:modality_compare} are aggregated over common operating conditions, including sunny, cloudy, and light-rain scenarios.

\begin{table}[h!]
\centering
\scriptsize
\captionsetup{justification=justified, singlelinecheck=false}
\caption{End-to-end tracking performance by modality and weather condition.}
\label{tab:modality_compare}
\setlength{\tabcolsep}{8pt}
\renewcommand{\arraystretch}{1.15}
\begin{tabular}{llccc}
\toprule
\textbf{Condition} & \textbf{Modality} & \textbf{MOTA}$\uparrow$ & \textbf{IDF1}$\uparrow$ \\
\midrule
\multirow{3}{*}{Sunny}
  & Camera-only & 67.7 & 73.6 \\
  & LiDAR-only  & 76.4 & 84.3 \\
  & \textbf{Fusion} & \textbf{78.6} & \textbf{86.0} \\
\midrule
\multirow{3}{*}{Cloudy}
  & Camera-only & 63.2 & 68.8 \\
  & LiDAR-only  & 60.1 & 66.9 \\
  & \textbf{Fusion} & \textbf{65.4} & \textbf{71.0} \\
\midrule
\multirow{3}{*}{Light rain}
  & Camera-only & 59.5 & 66.4 \\
  & \textbf{LiDAR-only} & \textbf{70.1} & \textbf{78.3} \\
  & Fusion & 65.8 & 71.2 \\
\bottomrule
\end{tabular}
\end{table}

In the sunny subset, the standalone camera tracker is more affected by occlusion, while LiDAR provides stronger geometric cues and more robust tracking, yielding a substantial gain over camera-only performance (MOTA: 76.4 vs. 67.7). Our fusion method further improves both MOTA (78.6) and IDF1 (86.0) by combining LiDAR geometry and motion consistency with camera-based semantic information.

In the cloudy subset, the camera performs comparatively well, whereas LiDAR under-detects small or partially occluded pedestrians. Fusion combines these complementary strengths, achieving the highest tracking accuracy (MOTA: 65.4) and identity consistency (IDF1: 71.0) among all modalities.

Under light rain, both sensors remain functional, though the camera experiences reduced contrast and occasional missed tracks while LiDAR performance degrades slightly due to rain-induced noise. Under these conditions, LiDAR-only achieves the best MOTA (70.1) and IDF1 (78.3), suggesting that condition-aware sensor selection may improve tracking performance.

\subsubsection{Real-Time Throughput and Scalability}
\begin{figure*}[t!]
\captionsetup{justification=justified, singlelinecheck=false}
\centering
\includegraphics[scale=0.16]{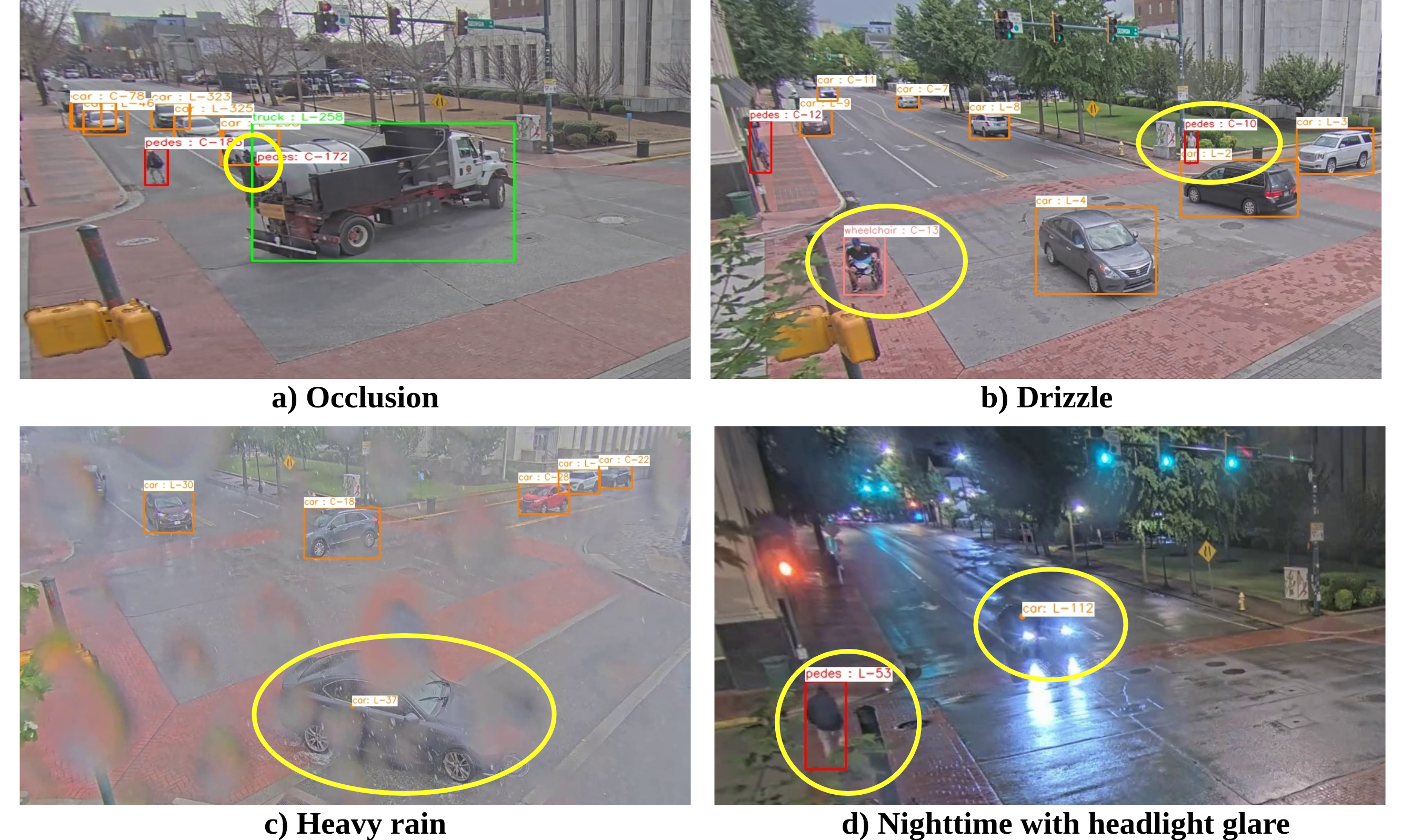}
\caption{Qualitative results demonstrating the robustness of our fusion method under challenging real-world scenarios. LiDAR-only detections are shown as projected ground-plane points, while camera and fused detections use bounding boxes. Object IDs are prefixed with \textbf{C} (camera) or \textbf{L} (LiDAR) to indicate the sensor that first detected each object.}
\label{fig:Result_2}
\end{figure*}

Table~\ref{tab:fps_results} summarizes component-wise throughput on two edge-computing platforms: the Jetson AGX Thor used in our deployment and the Jetson AGX Orin for comparison. LiDAR perception is performed by an external BlueCity unit and delivered to the edge node as an API stream at a fixed rate of 10\,Hz. Late fusion and multi-object tracking execute on the CPU, whereas calibration is run as an off-path process during system setup or occasional recalibration events due to its higher computational cost, with a single calibration pipeline active at any given time. Consequently, the steady-state end-to-end throughput of the system is effectively limited by the 10\,Hz LiDAR stream rather than by the camera perception or fusion components.

\begin{table}[h!]
\captionsetup{justification=justified, singlelinecheck=false}
\caption{Component throughput on AGX Thor and AGX Orin (FPS).}
\label{tab:fps_results}
\scriptsize
\centering
\setlength{\tabcolsep}{6pt}
\renewcommand{\arraystretch}{1.15}
\begin{tabular}{llcc}
\toprule
\textbf{Component} & \textbf{Method} & \textbf{AGX Thor} & \textbf{AGX Orin} \\
\midrule
Camera perception       & YOLOv11-m + ByteTrack         & 34.1 & 14.2 \\
LiDAR (offloaded)       & BlueCity API stream           & 10.0 & 10.0 \\
LiDAR (on-device)       & PointPillars + AB3DMOT        & 47.6 & 28.3 \\
Late fusion             & Distance-based (50 objs avg.) & 53.2 & 22.4 \\
\bottomrule
\end{tabular}
\end{table}

We define a \emph{CLIFE worker} as one complete CLIFE pipeline on the Jetson AGX Thor, covering camera perception, LiDAR API ingestion, and late-fusion processing. Since the camera worker (YOLOv11-m + ByteTrack) dominates GPU utilization, we focus scalability analysis on this component. A single camera worker achieves 34.1\,FPS in isolation, though the fused output is paced by the 10\,Hz LiDAR stream. As the number of parallel camera workers increases, per-worker throughput decreases due to GPU contention, but remains above 10\,Hz for up to five workers (20.6, 15.4, 12.4, and 10.4\,FPS for two through five workers) and drops to 8.8\,FPS with six workers.

As an alternative to the BlueCity backend, we evaluate a fully on-device LiDAR pipeline using PointPillars~\cite{lang2019pointpillars}, TensorRT via the NVIDIA LiDAR AI Solution~\cite{nvidia_lidar_ai_solution}, and AB3DMOT~\cite{weng2020ab3dmot}. The LiDAR pipeline alone achieves 47.6\,FPS with one worker and 13.5\,FPS with five parallel workers. In scalability tests excluding the additional $\sim$97\,ms delay for transferring LiDAR point clouds from the Ouster device, the on-device pipeline sustains four parallel CLIFE workers at 10.6\,FPS, one fewer than the BlueCity-offloaded setup.

Each deployed intersection hosts two cameras and two LiDAR units. A single Jetson AGX Thor sustains five parallel CLIFE workers at approximately 10\,FPS per stream, corresponding to two full intersections plus one additional camera-LiDAR stream, with larger deployments scaled by adding edge nodes in parallel.

\subsubsection{Qualitative Results}
\label{sec:Qualitative}
Fig.~\ref{fig:Result_2} presents qualitative results that highlight the robustness of our sensor fusion method in several challenging scenarios, as indicated by the yellow ellipses. These cases demonstrate the complementary strengths of the LiDAR and camera sensors:

\begin{itemize}
    \item \textbf{Occlusion}: The LiDAR sensor maintains a continuous track of a pedestrian who is temporarily occluded from the camera's view by a large truck.
    \item \textbf{Drizzle}: The camera's semantic capabilities enable the classification of specific VRU types, such as wheelchair users, and allow for the detection of pedestrians in the LiDAR sensor's blind spots.
    \item \textbf{Heavy rain}: During adverse weather like heavy rain, the performance of both sensors degrades significantly. Although this can cause frequent ID changes from individual sensors, our fusion mechanism compensates for detection failures by either sensor to ensure more continuous tracking.
    \item \textbf{Nighttime with headlight glare}: In conditions with poor lighting or glare, the camera's detection confidence decreases significantly, while LiDAR performance remains stable. For example, the camera may fail to detect a car due to glare, whereas the LiDAR sensor successfully maintains the detection. Our system also uses initial LiDAR detections of a partially occluded pedestrian (Fig. \ref{fig:Result_2}a) to maintain a track until the object is also clearly visible to the camera, at which point the tracks are fused.
\end{itemize}

These examples demonstrate continuous tracking across varied conditions and effective VRU classification using high-resolution imagery.
\section{Discussion and Future Work}

\ourpipeline's modular architecture supports long-term deployment by decoupling perception, calibration, and fusion, allowing individual components to be upgraded independently without re-engineering the full stack, reducing vendor lock-in and accelerating deployment cycles.

The distance-based late-fusion approach has notable limitations. Performance remains sensitive to extrinsic calibration accuracy despite the online refinement routine, and late-fusion propagates errors from upstream modules into the fused output. Reliance on a proprietary LiDAR API further constrains end-to-end optimization and scalability.

Future work will integrate trajectory prediction, risk assessment, and proactive alerting to shift from reactive perception toward preventive safety interventions. We plan to adapt the framework to public benchmarks~\cite{sekaran2025urbaning, mirlach2025r} and replace the commercial LiDAR backend with open-source alternatives for reproducible comparison. Incorporating cooperative V2X communication and multi-modal sensing remains a promising direction for improving robustness in occluded or structurally complex environments.
\section{Conclusion}

This paper presented the \ourpipeline framework, a real-time, edge-deployable system for robust roadside perception of vulnerable road users. By integrating modular perception components and a lightweight late-fusion strategy, our framework fuses complementary data from camera and LiDAR sensors to operate effectively in complex urban intersections. The system features a fully automatic, targetless calibration module designed as an on-demand, off-path process, which allows for remote recalibration without interrupting the primary real-time pipeline. This overall architecture facilitates scalable deployment with minimal human intervention.

Extensive field experiments conducted at a busy intersection demonstrated that the framework maintains reliable detection and tracking performance across diverse environmental and traffic conditions. Running on a Jetson AGX Thor, the system achieves real-time performance, with the core late-fusion module operating at 53.2 FPS. One edge node can support five parallel camera–LiDAR fusion pipelines at approximately 10 FPS per stream, corresponding to two full intersections plus one additional camera–LiDAR stream. CLIFE lays a practical foundation for next-generation roadside safety infrastructure that is intelligent, adaptive, and scalable across diverse urban environments.

{
    \small
    \bibliographystyle{IEEEtran}
    \bibliography{ref}
}

\end{document}